\documentclass[11pt,a4paper]{article}
\usepackage[hyperref]{naaclhlt2019}
\usepackage{times}
\usepackage{latexsym}
\usepackage{subfig}
\usepackage{graphicx}
\usepackage{url}
\usepackage{algorithm}
\usepackage{multirow}
\usepackage{booktabs}
\usepackage{tikz}
\usepackage{algpseudocode}
\usepackage{amsmath}
\usepackage{epsfig}
\usepackage{enumitem}
\setenumerate[1]{itemsep=0pt,partopsep=0pt,parsep=\parskip,topsep=5pt}
\setitemize[1]{itemsep=0pt,partopsep=0pt,parsep=\parskip,topsep=5pt}
\setdescription{itemsep=0pt,partopsep=0pt,parsep=\parskip,topsep=5pt}
\usepackage[font=small,labelfont=bf,textfont=md]{caption}

\aclfinalcopy 

\newcommand{\hide}[1]{} 

\title{Improving Distantly-supervised Entity Typing \\with Compact Latent Space Clustering}

\author{Bo Chen$^1$, Xiaotao Gu$^2$, Yufeng Hu$^1$, Siliang Tang$^1$\thanks{Corresponding Author.} , Guoping Hu$^3$, \\
    \textbf{Yueting Zhuang$^1$ \& Xiang Ren$^4$}\\
    $^1$Zhejiang University, $^2$University of Illinois at Urbana Champaign\\ 
    $^3$iFLYTEK Research, $^4$University of Southern California, \\
    \texttt{\{chenbo123, xiaofeem, siliang, yzhuang\}@zju.edu.cn},\\ \texttt{xiaotao2@illinois.edu}, \texttt{gphu@iflytek.com}, \texttt{xiangren@usc.edu}}
\date{}

\begin{document}
\maketitle

\begin{abstract}
Recently, distant supervision has gained great success on Fine-grained Entity Typing (FET). Despite its efficiency in reducing manual labeling efforts, it also brings the challenge of dealing with false entity type labels, as distant supervision assigns labels in a context-agnostic manner. Existing works alleviated this issue with partial-label loss, but usually suffer from confirmation bias, which means the classifier fit a pseudo data distribution given by itself. In this work, we propose to regularize distantly supervised models with Compact Latent Space Clustering (CLSC) to bypass this problem and effectively utilize noisy data yet.  Our proposed method first dynamically constructs a similarity graph of different entity mentions; infer the labels of noisy instances via label propagation. Based on the inferred labels, mention embeddings are updated accordingly to encourage entity mentions with close semantics to form a compact cluster in the embedding space, thus leading to better classification performance. Extensive experiments on standard benchmarks show that our CLSC model consistently outperforms state-of-the-art distantly supervised entity typing systems by a significant margin.
\end{abstract}

\section{Introduction}

Recent years have seen a surge of interests in fine-grained entity typing \textbf{(FET)} as it serves as an important cornerstone of several nature language processing tasks including relation extraction \cite{mintz2009distant}, entity linking \cite{DBLP:conf/aaai/RaimanR18}, and knowledge base completion \cite{dong2014knowledge}. To reduce manual efforts in labelling training data, distant supervision \cite{mintz2009distant} has been widely adopted by recent FET systems. With the help of an external knowledge base (KB), an entity mention is first linked to an existing entity in KB, and then labeled with all possible types of the KB entity as supervision. However, despite its efficiency, distant supervision also brings the challenge of \textbf{out-of-context noise}, as it assigns labels in a context agnostic manner. Early works usually ignore such noise in supervision \cite{ling2012fine,shimaoka2016attentive}, which dampens the performance of distantly supervised models. 
\begin{figure}[t]
    \centering
    \hspace{-0.7em}
    \subfloat{
    \includegraphics[width=0.5\linewidth]{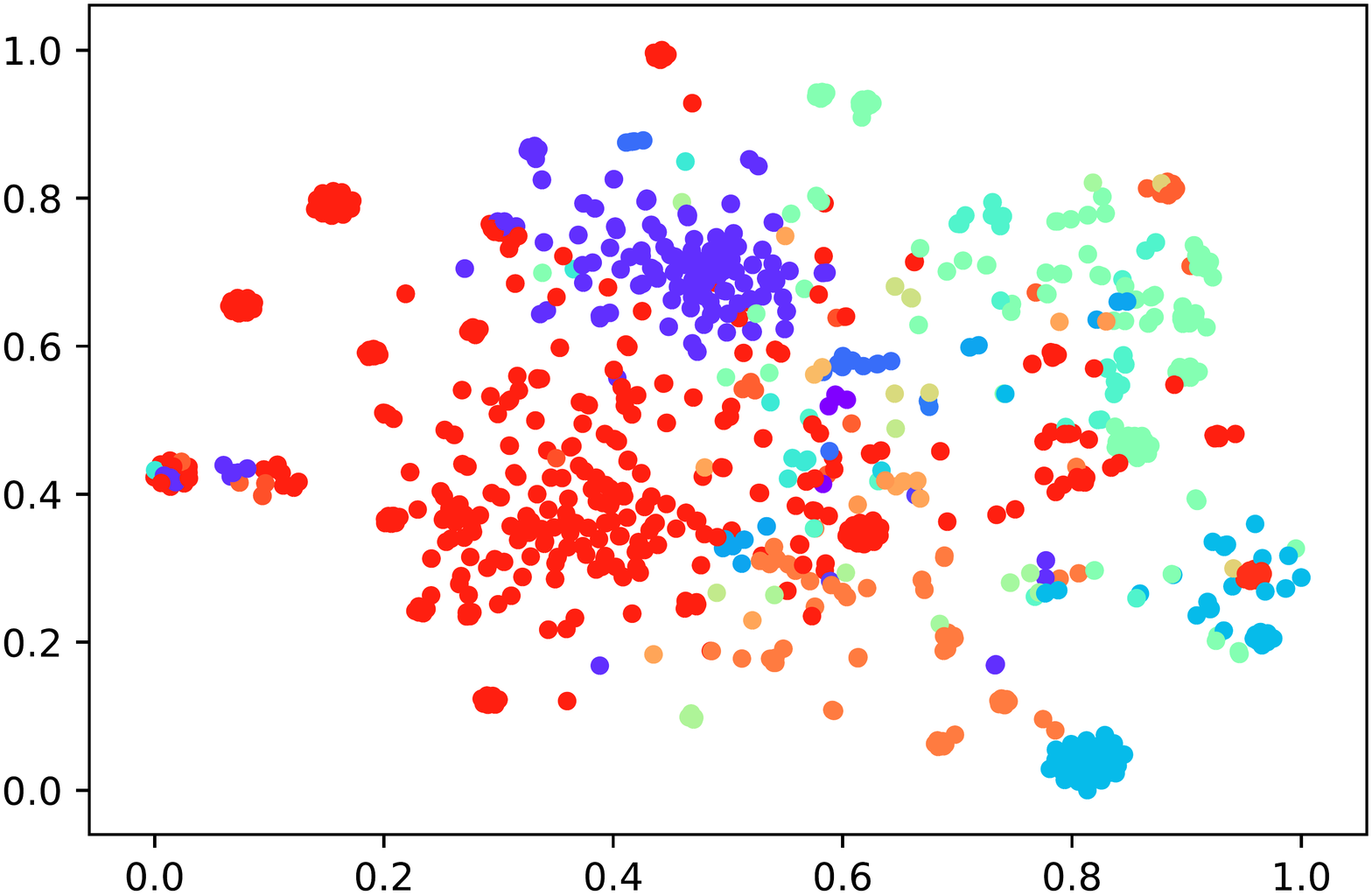}}
    \subfloat{
    \includegraphics[width=0.5\linewidth]{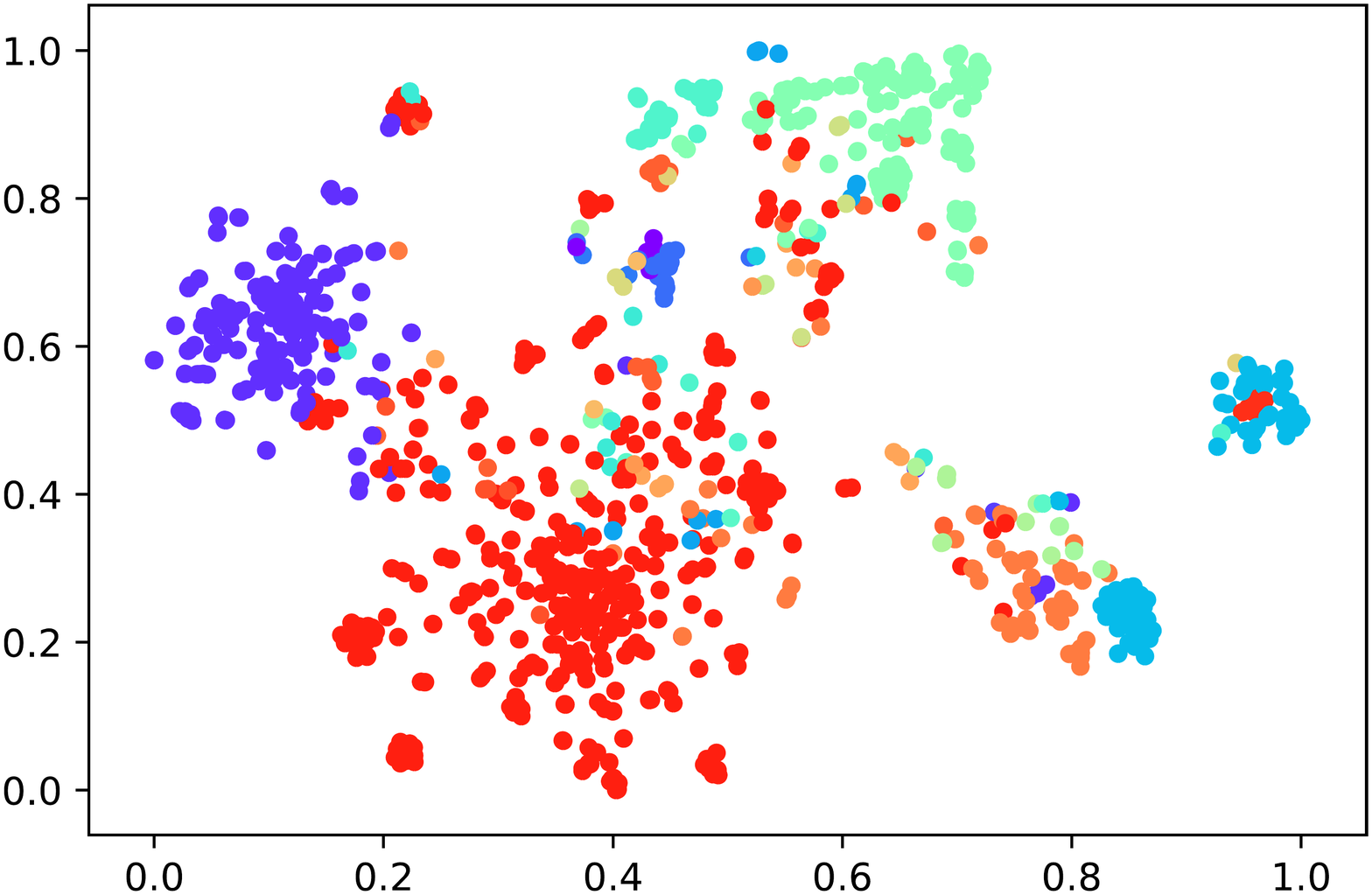}}
    \caption{T-SNE visualization of the mention embeddings generated by NFETC (left) and CLSC (right) on the BBN dataset. Our model (CLSC) clearly groups mentions of the same type into a compact cluster.}
    \label{fig:tsneVis}
    \vspace{-1em}
\end{figure}
\begin{figure*}
	\includegraphics[width=1.0\linewidth,height=0.4\linewidth]{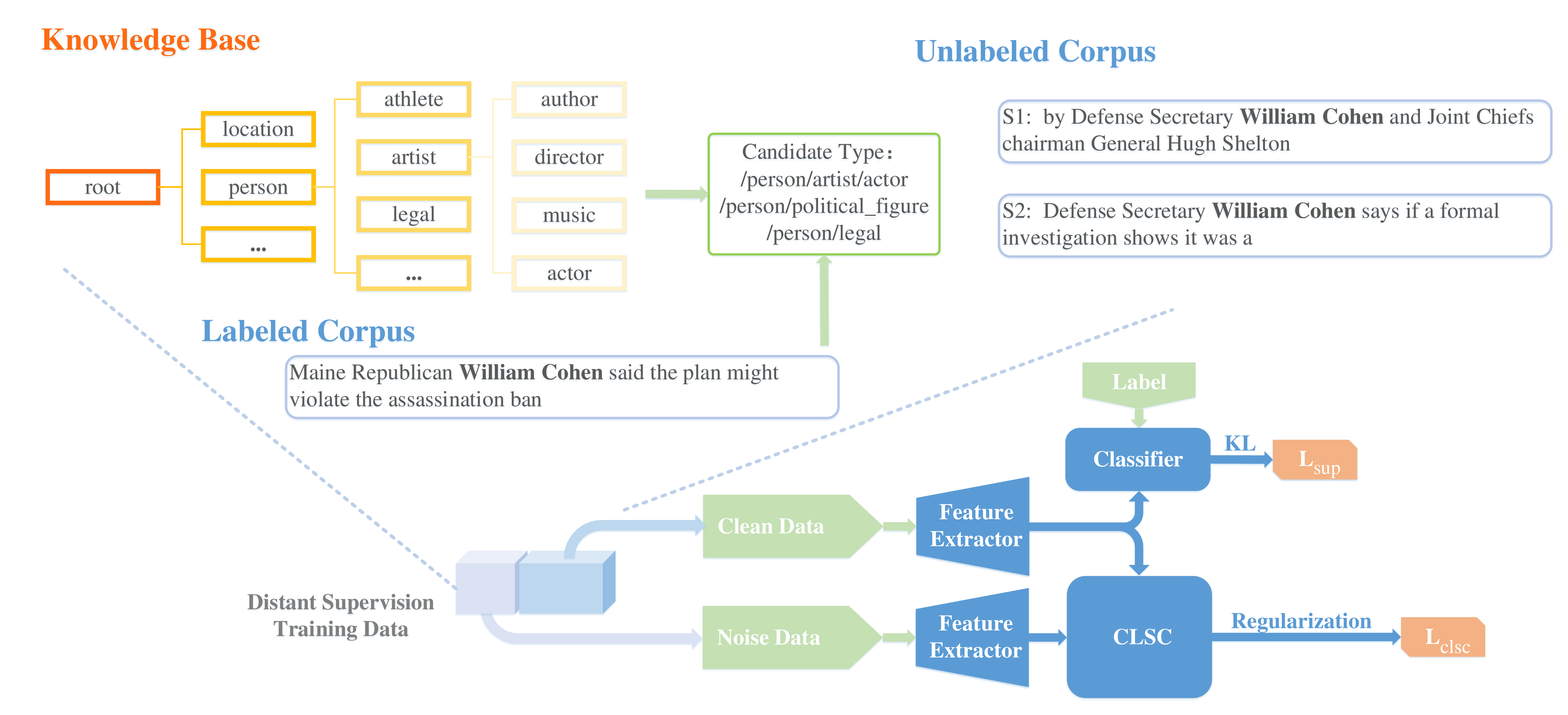}
	\caption{The overall framework of CLSC. We calculate classification loss only on clean data, while regularize the feature extractor with CLSC using both clean and noisy data.}
	\label{fig:fk}
\end{figure*}

Towards overcoming out-of-context noise, two lines of work have been proposed to distantly supervised FET. The first kind of work try to filter out noisy labels using heuristic rules~ \cite{gillick2014context}.
However, such heuristic pruning significantly reduces the amount of training data, and thus cannot make full use of distantly annotated data. 
In contrast, the other thread of works try to incorporate such imperfect annotation by partial-label loss (\textbf{PLL}). 
The basic assumption is that, \emph{for a noisy mention, the maximum score associated with its candidate types should be greater than the scores associated with any other non-candidate types} \cite{ren2016afet,abhishek2017fine,xu2018neural}.
Despite their success, \textbf{PLL}-based models still suffer from \textbf{\emph{Confirmation Bias}} by taking its own prediction as optimization objective in the next step. 
Specifically, given an entity mention, if the typing system selected a wrong type with the maximum score among all candidates, it will try to further maximize the score of the wrong type in following optimization epoches (in order to minimize \textbf{PLL}), thus amplifying the confirmation bias. 
Such bias starts from the early stage of training, when the typing model is still very suboptimal, and can accumulate in training process. Related discussion can be also found in the setting of semi-supervised learning \cite{lee2006fine,Laine2017iclr,tarvainen2017mean}.

In this paper, we propose a new method for distantly supervised fine-grained entity typing. Enlightened by \cite{pmlr-v80-kamnitsas18a}, we propose to effectively utilize imperfect annotation as model regularization via \textbf{C}ompact \textbf{L}atent \textbf{S}pace \textbf{C}lustering \textbf{(CLSC)}. More specifically, our model encourages the feature extractor to group mentions of the same type as a compact cluster (dense region) in the representation space, which leads to better classification performance. For training data with noisy labels, instead of generating pseudo supervision by the typing model itself, we dynamically construct a similarity-weighted graph between clean and noisy mentions, and apply label propagation on the graph to help the formation of compact clusters. Figure \ref{fig:tsneVis} demonstrates the effectiveness of our method in clustering mentions of different types into dense regions. In contrast to \textbf{PLL}-based models, we do not force the model to fit pseudo supervision generated by itself, but only use noisy data as part of regularization for our feature extractor layer, thus avoiding  bias accumulation. \\
Extensive experiments on standard benchmarks show that our method consistently outperforms state-of-the-art models. Further study reveals that, the advantage of our model over the competitors gets even more significant as the portion of noisy data rises.

\begin{figure*}
    \centering
	\includegraphics[width=1.05\linewidth,height=0.4\linewidth]{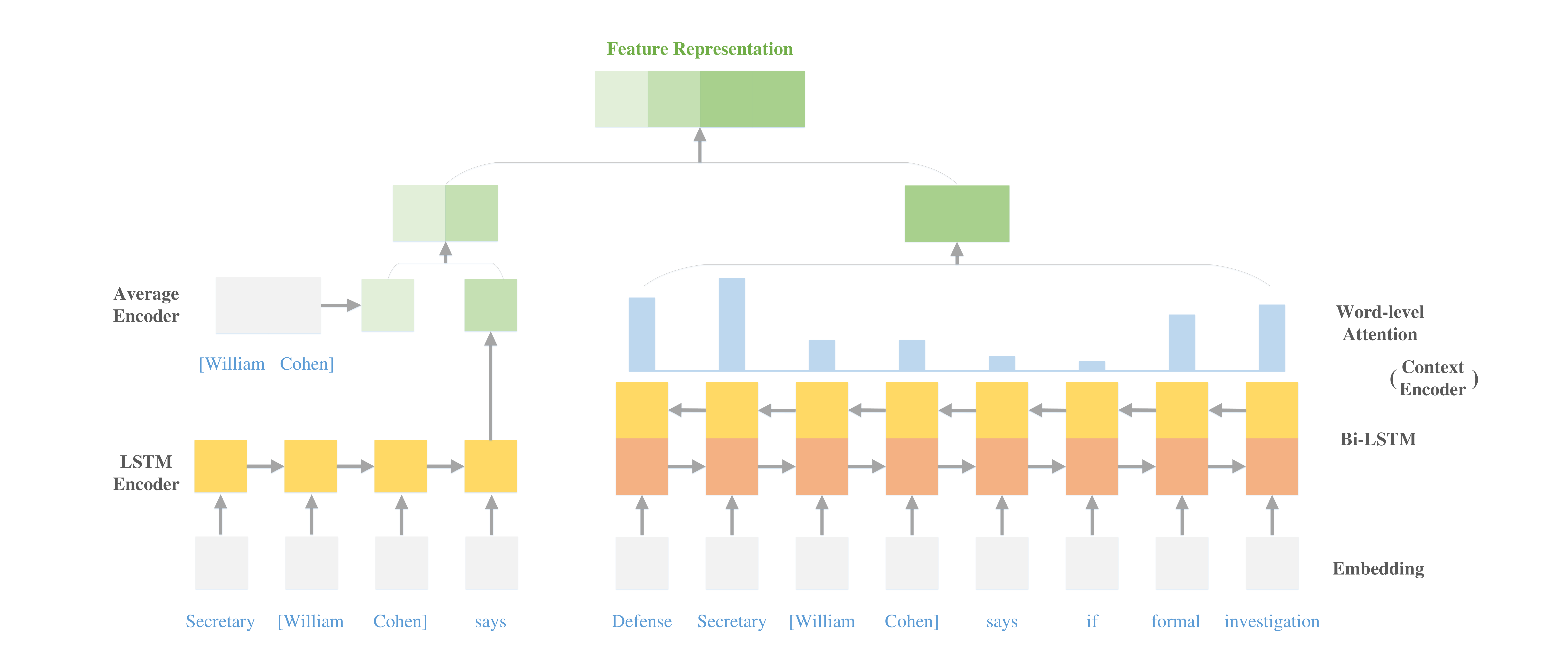}
	\caption{The architecture of feature extractor  $z((m_i,c_i);\theta_z)$}
	\label{fig:feature extractor}
\end{figure*}
\section{Problem Definition}
Fine-grained entity typing takes a corpus and an external knowledge base (KB) with a type hierarchy $\mathcal{Y}$ as input. Given an entity mention (i.e., a sequence of token spans representing an entity) in the corpus, our task is to uncover its corresponding type-path in $\mathcal{Y}$ based on the context. 

By applying distant supervision, each mention is first linked to an existing entity in  KB, and then labeled with all its possible types. Formally, a labeled corpus can be represented as triples $\mathcal{D}=\{(m_i,c_i,\mathcal{Y}_i)\}_{i=1}^n$, where $m_i$ is the $i$-th mention, $c_i$ is the context of $m_i$, $\mathcal{Y}_i$ is the set of candidate types of $m_i$. Note that types in $\mathcal{Y}_i$ can form one or more type paths. In addition, we denote  all terminal (leaf) types of each type path in $\mathcal{Y}_i$ as the target type set $\mathcal{Y}_i^t$ (\textit{e.g.}, for $\mathcal{Y}_i=\{artist, teacher,person\}$, $\mathcal{Y}_i^t=\{artist,teacher\}$). This setting is also adopted by \cite{xu2018neural}.

As each entity in KB can have several type paths, \emph{out-of-context} noise may exist when $\mathcal{Y}_i$ contains type paths that are irrelevant to $m_i$ in context $c_i$. In this work, we argue triples where $\mathcal{Y}_i$ contains only one type path (i.e., $|\mathcal{Y}^t_i| = 1$) as \textbf{clean data}. Other triples are treated as \textbf{noisy data}, where $\mathcal{Y}_i$ contains both the true type path and irrelevant type paths. 
Noisy data usually takes a considerable portion of the entire dataset.
The major challenge for distantly supervised typing systems is to incorporate both clean and noisy data to train high-quality type classifiers.

\section{The Proposed Approach}
\label{sec:method}
\noindent
\textbf{Overview.}
The basic assumptions of our idea are: (1) all mentions belong to the same type should be close to each other in the representation space because they should have similar context, (2) similar contexts lead to the same type. For clean data, we compact the representation space of the same type to comply (1). For noisy data, given assumption (2), we infer the their type distributions via label propagation and candidate types constrain. \\
Figure \ref{fig:fk} shows the overall framework of the proposed method. Clean data is used to train classifier and feature extractor end-to-endly, while noisy data is only used in CLSC regularization. Formally,
given a batch of samples $\{(m_i,c_i,\mathcal{Y}_i^t)\}_{i=1}^B$, we first convert each sample $(m_i,c_i)$ into a real-valued vector $z_i$ via a feature extractor $z((m_i,c_i);\theta_z)$ parameterized by $\theta_z$. Then a type classifier $g(z_i;\theta_g)$ parameterized by $\theta_g$ gives the posterior $P(y|z_i;\theta_g)$. By incorporating CLSC regularization in the objective function, we encourage the feature extractor $z$ to group mentions of the same type into a compact cluster, which facilitates classification as is shown in Figure \ref{fig:tsneVis}. Noisy data enhances the formation of compact clusters with the help of label propagation.

\subsection{Feature Extractor}

Figure \ref{fig:feature extractor} illustrates our feature extractor. For fair comparison, we adopt the same feature extraction pipeline as used in \cite{xu2018neural}. The feature extractor is composed of an embedding layer and two encoders which encode mentions and contexts respectively.\\
\smallskip
\noindent
\textbf{Embedding Layer:}\label{sec:emb layer}
The output of this layer is a concatenation of word embedding and word position embedding. We use the popular 300-dimensional word embedding supplied by \cite{pennington2014glove} to capture the semantic information and random initialized position embedding \cite{zeng2014relation} to acquire information about the relation between words and the mentions.

Formally, Given a
word embedding matrix $W_{word}$ of shape $d_w\times|V|$,
where $V$ is the vocabulary and $d_w$ is the size of word embedding, each column of $W_{word}$ represents a specific word $w$ in $V$. We map each word $w_j$ in $(m_i,c_i)$ to a word embedding $\mathbf{{w}_j^d}\in R^{d_w}$. Analogously, we get the word position embedding $\mathbf{w}_j^p\in R^{d_p}$ of each word according to the relative distance between the word and the mention, we only use a fixed length context here. The final embedding of the j-th word is $\bf{w}_j^E= [\bf{{w}_j^d},\bf{{w}_j^p}]$. \\

\smallskip
\noindent
\textbf{Mention Encoder:}
To capture lexical level information of mentions, an averaging mention encoder and a LSTM mention encoder \cite{hochreiter1997long} is applied to encode mentions. Given $m_i=(w_s,w_{s+1},\cdots,w_{e})$, the averaging mention representation $r_{a_i}\in R^{d_w}$ is :
\begin{equation}
    r_{a_i}=\frac{1}{e-s+1}\sum_{j=s}^e\bf{{w}_j^d}
\end{equation}
By applying a LSTM over an extended mention $(w_{s-1},w_s,w_{s+1},\cdots,w_e,w_{e+1})$, we get a sequence  $(h_{s-1},h_s,h_{s+1},\cdots,h_e,h_{e+1})$. We use $h_{e+1}$ as LSTM mention representation $r_{l_i}\in R^{d_l}$. The final mention representation is $r_{m_i}=[r_{a_i},r_{l_i}]\in R^{{d_w}+{d_l}}$.\\
\smallskip
\noindent
\textbf{Context Encoder:}
A bidirectional LSTM with $d_l$ hidden units is  employed to encode embedding sequence $(\bf{w}_{s-W}^E,\bf{w}_{s-W+1}^E,\cdots,\bf{w}_{e+W}^E)$:
\begin{equation}
\begin{aligned}
    \overrightarrow{h_{j}}=& LSTM(\overrightarrow{h_{j-1}},\bf{w}_{j-1}^E)\\
    \overleftarrow{h_{j}}=& LSTM(\overleftarrow{h_{j-1}},\bf{w}_{j-1}^E)\\
    h_{j}=&[\overrightarrow{h_{j}}\oplus\overleftarrow{h_{j}}]
\end{aligned}
\end{equation}
where $\oplus$ denotes element-wise plus. Then, the word-level attention mechanism computes a score $\beta_{i,j}$ over different word $j$ in the context $c_i$ to get the final context representation $r_{c_i}$:
\begin{equation}
\begin{aligned}
    \alpha_j=& w^Ttanh(h_j)\\
    \beta_{i,j}=&\frac{exp(\alpha_j)}{\sum\limits_{k}exp(\alpha_k)}\\
    r_{c_i}=&\sum\limits_{j}\beta_{i,j}h_{i,j}
\end{aligned}
\end{equation}
We use $r_i=[r_{m_i},r_{c_i}]\in R^{d_z}=R^{{d_w}+{d_l}+{d_l}}$ as the feature representation of $(m_i,c_i)$ and use a Neural Networks $q$ over $r_i$ to get the feature vector $z_i$. $q$ has $n$ layers with $h_n$ hidden units and use ReLu activation.

\subsection{Compact Latent Space Clustering for Distant Supervision}
\label{sec: CSLC}
\begin{figure*}
	\includegraphics[width=1.0\linewidth,height=0.32\linewidth]{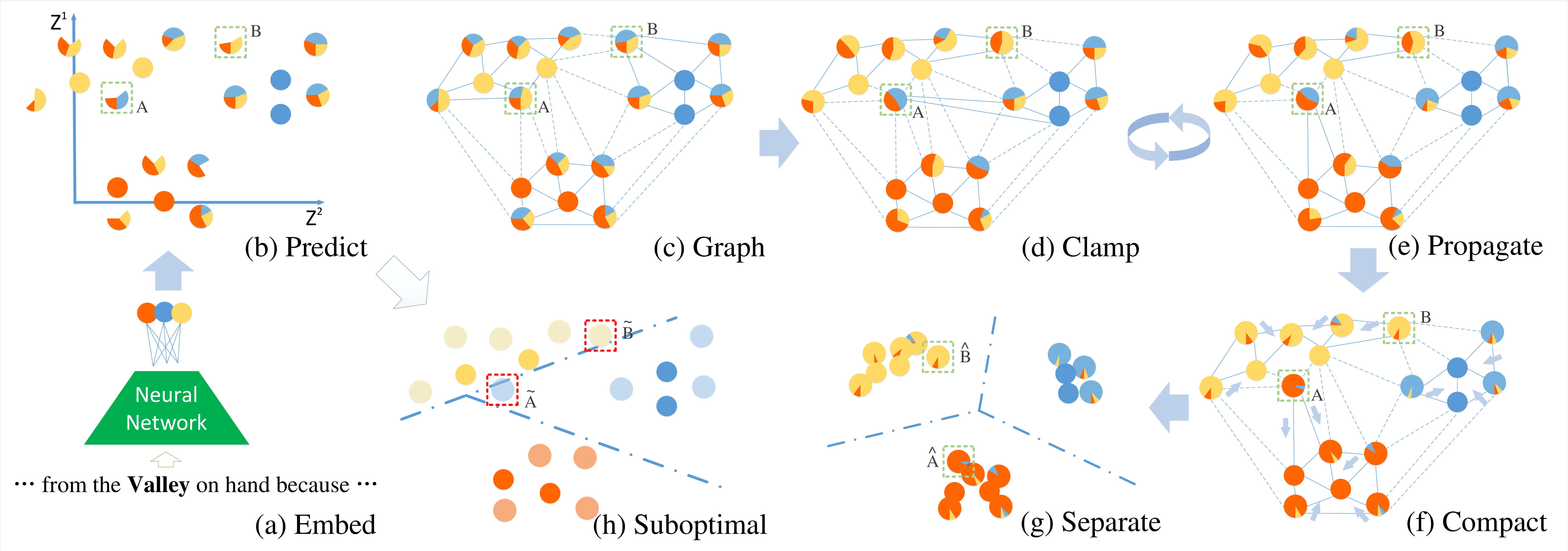}

    \caption{A demonstration of the CLSC process. (a) represents the feature extraction step; (b)$\rightarrow$(h) shows the traditional type classification process (each color represents one candidate type), where suboptimal classifiers make predictions for each mention and misclassifies A into the Blue type; (c)$\rightarrow$(d)$\rightarrow$(e)$\rightarrow$(f)$\rightarrow$(g) demonstrates the process of CLSC as described in Section \ref{sec:method}. Through label propagation and compact clustering, our model is able to group mentions of the same type into a dense region and leaves clear separation boundaries in sparse regions.}
	\label{fig:CLSC}
\end{figure*}
The overview of CLSC regularization is exhibited
in Figure \ref{fig:CLSC}, which includes three steps: dynamic graph construction (Figure \ref{fig:CLSC}c), label propagation (Figure \ref{fig:CLSC}d, e) and Markov chains (Figure \ref{fig:CLSC}g). The idea of compact clustering for semi-supervised learning is first proposed by \cite{pmlr-v80-kamnitsas18a}. The basic idea is to encourage mentions of the same type to be clustered into a dense region in the embedding space. We introduce more details of CLSC for distantly supervised FET in following sections.

\smallskip
\noindent
\textbf{Dynamic Graph Construction:}
We start by creating a fully connected graph $G$ over the batch of samples $\mathbf{Z}=\{z_i\}_{i=1}^B$, as shown in Figure \ref{fig:CLSC}c\footnote{$\mathbf{Z}=\{z_i\}_{i=1}^B$ is a small subsample of the entire data, we didn't observe significant performance gain when the batch size increases.}. Each node of $G$ is a feature representation $z_i$, while the distance between nodes is defined by a scaled dot-product distance function \cite{vaswani2017attention}:
\begin{equation}
\begin{aligned}
    A_{ij}=&exp(\frac{z_i^Tz_j}{\sqrt{d_z}}), \forall z_i,z_j\in{\mathbf{Z}}\\
    A=&exp(\frac{Z^TZ}{\sqrt{d_z}})\label{eq:distancefunc}
\end{aligned}
\end{equation}
Each entry $A_{ij}$ measures the similarity between $z_i$ and $z_j$, $A\in{R}^{B\times{B}}$ can be viewed as the weighted adjacency matrix of $G$.

\smallskip
\noindent
\textbf{Label Propagation:}\label{subsub:lp}
The end goal of CLSC is to cluster mentions of the same type to a dense region. For mentions which have more than one labeled types, we apply label propagation (\textbf{LP}) on $G$ to estimate their type distribution. Formally, we denote $\mathbf{\Phi}\in R^{B\times{K}}$ as the label propagation posterior of a training batch. 

The original label propagation proposed by \cite{zhu2002learning} uses a transition matrix $H$ to model the probability of a node $i$ propagating its type posterior $\mathbf{\phi}_i=P(y_i|x_i)\in{R}^K$ to the other nodes.
Each entry of the transition matrix $H\in{R}^{B\times{B}}$ is defined as:
\begin{equation}
    H_{ij}=A_{ij}/\sum_b{A_{ib}}
\end{equation}
The original label propagation algorithm is defined as:
\begin{enumerate}
    \item Propagate the label by transition matrix $H$, $\mathbf{\Phi}^{(t+1)}={H}\mathbf{\Phi}^{(t)}$
    \item Clamp the labeled data to their true labels. Repeat from step 1 until $\mathbf{\Phi}$ converges
\end{enumerate}
In this work $\Phi^{(0)}$ is randomly initialized\footnote{We also explored other initialization (e.g. uniform initialization), but found no essential performance difference between different initialization setups.}.
Unlike unlabeled data in semi-supervised learning, distantly labeled mentions in FET have a limited set of candidate types. Based on this observation, 
We assume that $(m_i,c_i)$ can only transmit and receive probability of types in $\mathcal{Y}_i^t$ no matter it is noisy data or clean data. Formally,
define a ${B}\times{K}$ indicator matrix $M\in{R}^{B\times{K}}$, where $M_{ij}=1$ if type j in $\mathcal{Y}_i^t$ otherwise $0$, where $B$ is the batch size and $K$ is the number of types. Our clamping step relies on $M$ as is shown in Figure \ref{fig:CLSC}d:
\begin{equation}
    \Phi_{ij}^{(t+1)}\leftarrow\Phi_{ij}^{(t+1)}M_{ij}/\sum_k{\Phi_{ik}^{(t+1)}M_{ik}}
\end{equation}
For convenience, we iterate through these two steps $S_{lp}$ times, $S_{lp}$ is a hyperparameter.

\smallskip
\noindent
\textbf{Compact Clustering:}
\label{sec:method-compact}
The \textbf{LP} posterior $\mathbf{\Phi}=\mathbf{\Phi}^{(S_{lp}+1)}$ is used to judge the label agreement between samples. In the desired optimal state, transition probabilities between samples should be uniform inside the same class, while be zero between different classes.
Based on this assumption, the desirable transition matrix $T\in{R}^{B\times{B}}$ is defined as:
\begin{equation}
    T_{ij}=\sum_{k=1}^K\Phi_{ik}\frac{\Phi_{jk}}{m_k}, m_k=\sum_{b=1}^B\Phi_{bk}
\end{equation}
$m_k$ is a normalization term for class $k$.
Transition matrix $H$ derived from $z((m_i,c_i);\theta_z)$ should be in keeping with $T$. Thus we minimize the cross entropy between $T$ and $H$:
\begin{equation}
    \mathcal{L}_{1-step}=-\frac{1}{B^2}\sum_{i=1}^B\sum_{j=1}^BT_{ij}log(H_{ij})\label{1-step loss}
\end{equation}
\noindent For instance, if $T_{ij}$ is close to 1, $H_{ij}$ needs to be bigger, which results in the growth of $A_{ij}$ and finally optimize $\theta_z$ (Eq.\ref{eq:distancefunc}). 
The loss $\mathcal{L}_{1-step}$ has largely described the regularization we use in $z((m_i,c_i);\theta_z)$ for compression clustering.

In order to keep the structure of existing clusters, \cite{pmlr-v80-kamnitsas18a} proposed an extension of $\mathcal{L}_{1-step}$ to the case of \textbf{Markov chains} with multiple transitions between samples, which should remain within a single class. The extension maximizes probability of paths that only traverse among samples belong to one class.
Define $E\in{R}_{B\times{B}}$ as:
\begin{equation}
    E=\mathbf{\Phi}^T\mathbf{\Phi}
\end{equation}
$E_{ij}$ measures the label similarities between $z_i$ and $z_j$, which is used to mask the transition between different clusters. The extension is given by:
\begin{equation}
\begin{aligned}
    H^{(1)} = &H\\
    H^{(s)} = &(H\odot{E})^{(s-1)}H\\
    = &(H\odot{E})H^{(s-1)},
\end{aligned}
\end{equation}
where $\odot{}$ is Hadamard Product, and $H_{ij}^{(s)}$ is the probability of a Markov process to transit from node $i$ to node $j$ after $s-1$ steps within the same class. The extended loss function models paths of different length $s$ between samples on the graph:
\begin{equation}
    \mathcal{L}_{clsc}=-\frac{1}{S_m}\frac{1}{B^2}\sum_{s=1}^{S_m}\sum_{i=1}^B\sum_{j=1}^BT_{ij}log(H_{ij}^{(s)}).\label{eq:cclploss}
\end{equation}
For $S_m=1$, $\mathcal{L}_{clsc}=\mathcal{L}_{1-step}$. By minimizing the cross entropy between $T$ and $H^{(s)}$ (Eq.\ref{eq:cclploss}), $\mathcal{L}_{clsc}$ compact paths of different length between samples within the same class. Here, $S_{m}$ is a hyper-parameter to control the maximum length of Markov chain. $\mathcal{L}_{clsc}$ is added to the final objective function as regularization to encourage compact cluttering.

\subsection{Overall Objective}
Given the representation of a mention, the type posterior is given by a standard softmax classifier parameterized by $\theta_g$:
\begin{equation}
    P(\hat{y_i}|z_i;\theta_g)=softmax(W_c{z_i}+b_c),
\end{equation}
where $W_c\in{R^{K\times{d_z}}}$ is a parameter matrix, $b\in{R^K}$ is the bias vector, where $K$ is the number of types. The predicted type is then given by $\hat{t_i}=argmax_{y_i}P(\hat{y_i}|z_i;\theta_g)$.

Our loss function consists of two parts. $\mathcal{L}_{sup}$ is supervision loss defined by KL divergence:
\begin{equation}
\begin{aligned}
    L_{sup}=&-\frac{1}{B_c}\sum_{i=1}^{B_c}\sum_{k=1}^Ky_{ik}log(P(y_i|z_i;\theta_g))_k \label{eq:clean loss}
\end{aligned}
\end{equation}
Here $B_c$ is the number of clean data in a training batch, K is the number of target types. The regularization term is given by $\mathcal{L}_{clsc}$. Hence, the overall loss function is:
\begin{equation}
    \mathcal{L}_{final}=\mathcal{L}_{sup}+\lambda_{clsc}\times\mathcal{L}_{clsc}
\end{equation}
$\lambda_{clsc}$ is a hyper parameter to control the influence of CLSC.

\begin{table*}[]
\small
\centering
\begin{tabular}{llcccccc}
\toprule[2pt]
\multicolumn{2}{c}{\multirow{2}{*}{\textbf{Method}}}             & \multicolumn{3}{c}{\textbf{OntoNotes}}                                                                                   & \multicolumn{3}{c}{\textbf{BBN}}                                                                                        \\ \cmidrule[1pt]{3-8} 
\multicolumn{2}{c}{}                                             & \textbf{Strict Acc.}                   & \multicolumn{1}{l}{\textbf{Macro F1}}  & \multicolumn{1}{l}{\textbf{Micro F1}} & \textbf{Strict Acc.}                   & \textbf{Macro F1}                      & \textbf{Micro F1}                      \\ \midrule[1pt]
\multicolumn{2}{l}{\textbf{AFET} \cite{ren2016afet}}      & 55.3                                   & 71.2                                   & 64.6                                   & 68.3                                   & 74.4                                   & 74.7                                   \\
\multicolumn{2}{l}{\textbf{{AAA}} \cite{abhishek2017fine}}       & 52.2                                   & 68.5                                   & 63.3                                   & 65.5                                   & 73.6                                   & 75.2                                   \\
\multicolumn{2}{l}{\textbf{Attentive}~ \cite{shimaoka2016attentive}} & 51.7                                  & 71.0                                  & 64.91                                  & 48.4                                   & 73.2                                   & 72.4                                   \\
\multicolumn{2}{l}{\textbf{PLE+HYENA} \cite{ren2016label}} & 54.6                                   & 69.2                                   & 62.5                                   & 69.2                                   & 73.1                                   & 73.2                                   \\
\multicolumn{2}{l}{\textbf{{PLE+FIGER}}~~ \cite{ren2016label}} & 57.2                                   & 71.5                                   & 66.1                                   & 68.5                                   & 77.7                                   & 75.0                                   \\ \midrule[1pt] \midrule[1pt]
\multirow{2}{*}{\textbf{NFETC}}                 & $clean$        & 54.4$\pm$0.3                           & 71.5$\pm$0.4                           & 64.9$\pm$0.3                           & 71.2$\pm$0.2                           & 77.1$\pm$0.3                           & 76.9$\pm$0.3                           \\ \cmidrule[1pt]{2-8} 
                                                & $+noisy$       & 54.8$\pm$0.4                           & 71.8$\pm$0.4                           & 65.0$\pm$0.4                           & 73.8$\pm$0.6                           & 78.4$\pm$0.6                           & 78.9$\pm$0.6                           \\ \midrule[1pt]
\multirow{2}{*}{\textbf{NFETC}\textsubscript{$hier$}}            & $clean$        & 59.6$\pm$0.2                           & 76.1$\pm$0.2                           & 69.7$\pm$0.2                           & 70.3$\pm$0.3                           & 76.8$\pm$0.3                           & 76.6$\pm$0.2                           \\ \cmidrule[1pt]{2-8} 
                                                & $+noisy$       & 60.2$\pm$0.2                           & 76.4$\pm$0.1                           & 70.2$\pm$0.2                           & \textbf{73.9$\pm$1.2}                 & 78.8$\pm$1.2 & \textbf{79.4$\pm$1.1} \\ \midrule[1pt]\midrule[1pt]
\multirow{2}{*}{\textbf{NFETC-CLSC}}            & $clean$        & 59.1$\pm$0.4                           & 75.3$\pm$0.3                           & 69.1$\pm$0.3                           & 73.0$\pm$0.3                           & 79.0$\pm$0.3                           & 78.8$\pm$0.3                           \\ \cmidrule[1pt]{2-8} 
                                                & $+noisy$       & 59.6$\pm$0.2                           & 75.5$\pm$0.4                           & 69.3$\pm$0.4                           & \textbf{74.7$\pm$0.3} & \textbf{80.7$\pm$0.2} & \textbf{80.5$\pm$0.2} \\ \midrule[1pt]
\multirow{2}{*}{\textbf{NFETC-CLSC}\textsubscript{$hier$}}       & $clean$        & 61.5$\pm$0.3                           & 77.4$\pm$0.3                           & 71.4$\pm$0.4                           & 70.5$\pm$0.2                           & 78.2$\pm$0.2                           & 78.0$\pm$0.2                           \\ \cmidrule[1pt]{2-8} 
                                                & $+noisy$       & \textbf{62.8$\pm$0.3} & \textbf{77.8$\pm$0.4} & \textbf{72.0$\pm$0.4} & 71.9$\pm$0.3                           & 79.8$\pm$0.4                           & 79.5$\pm$0.3                           \\ \bottomrule[2pt]
\end{tabular}
\caption{Performance comparision of FET systems on the two datasets.}\label{tb:results}
\end{table*}

\section{Experiments}
\begin{table}[]
\centering
\small
\begin{tabular}{|l|l|l|}
\hline
                  & \textbf{OntoNotes} & \textbf{BBN}   \\ \hline
\textbf{\#types}                   & 89        & 47    \\ \hline
\textbf{Max hierarchy depth}       & 3         & 2     \\ \hline
\textbf{\#mentions-train}       & 253241    & 86078 \\ \hline
\textbf{\#mentions-test}        & 8963      & 12845 \\ \hline
\textbf{\%clean mentions-train} & 73.13     & 75.92 \\ \hline
\textbf{\%clean mentions-test}     & 94.00     & 100   \\ \hline
\textbf{Average $|\mathcal{Y}_i^t|$} & 1.40      & 1.26  \\ \hline
\end{tabular}
\caption{Detailed statistics of the two datasets.}
\label{tb:stati}
\end{table}
\subsection{Dataset}
We evaluate our method on two standard benchmarks: OntoNotes and BBN:
\begin{itemize}
    \item \textbf{OntoNotes:} The OntoNotes dataset is composed of sentences from the Newswire part of OntoNotes corpus \cite{weischedel2013ontonotes}. \cite{gillick2014context} annotated the training part with the aid of DBpedia spotlight \cite{daiber2013improving}, while the test data is manually annotated.
    \item \textbf{BBN:} The BBN dataset is composed of sentences from Wall Street Journal articles and is manually annotated by \cite{weischedel2005bbn}. \cite{ren2016afet} regenerated the training corpus via distant supervision.
\end{itemize}
In this work we use the preprocessed datasets provided by \cite{abhishek2017fine,xu2018neural}.  Table \ref{tb:stati} shows detailed statistics of the datasets.

\subsection{Compared Methods}
We compare the proposed method with several state-of-the-art FET systems\footnote{The baselines result are reported
on \cite{abhishek2017fine,xu2018neural} in addition to performance of NFETC on BBN, we search the hyper parameters for it. \cite{xu2018neural} didn't report the results on BBN}: 
\begin{itemize}
    \item \textbf{Attentive} \cite{shimaoka2016attentive} uses an attention based feature extractor and  doesn't distinguish clean from noisy data;
    \item \textbf{AFET} \cite{ren2016afet} trains label embedding with partial label loss;
    \item \textbf{AAA} \cite{abhishek2017fine} learns joint representation of mentions and type labels;
    \item \textbf{PLE+HYENA/FIGER} \cite{ren2016label} proposes heterogeneous partial-label embedding for label noise reduction to boost typing systems. We compare two PLE models with HYENA \cite{yogatama2015embedding} and FIGER \cite{ling2012fine} as the base typing system respectively;
    \item \textbf{NFETC} \cite{xu2018neural} trains neural fine-grained typing system with hierarchy-aware loss. We compare the performance of the NFETC model with two different loss functions: partial-label loss and \textbf{PLL}+hierarchical loss. We denote the two variants as $\mathbf{NFETC}$ and $\mathbf{NFETC}_{hier}$ respectively;
    \item \textbf{NFETC-CLSC} is the proposed model in this work. We use the NFETC model as our base model, based on which we apply Compact Latent Space Clustering Regularization as described in Section \ref{sec: CSLC}; Similarly, we report results produced by using both KL-divergense-based loss ($\textbf{NFETC-}\mathbf{CLSC}$) and \textbf{KL}+hierarchical loss ($\textbf{NFETC-}\mathbf{CLSC}_{hier}$).
\end{itemize}
\subsection{Evaluation Settings}
For evaluation metrics, we adopt strict accuracy, loose macro, and loose micro F-scores widely used in the FET task \cite{ling2012fine}.
To fine tuning the hyper-parameters, we randomly sampled 10\% of the test set as a development set for both datasets. With the fine-tuned hyper-parameter as mentioned in \ref{sec:hp}, we run the model five times and report the average strict accuracy, macro F1 and micro F1 on the test set. 

\subsection{Hyper Parameters}\label{sec:hp}
We search the hyper parameter of Ontonotes and BBN respectively via Hyperopt proposed by \cite{bergstra2013hyperopt}. Hyper parameters are shown in \textbf{Appendix A}.
We optimize the model via Adam Optimizer. The full hyper parameters includes the learning rate $lr$, the dimension $d_p$ of word position embedding, the dimension $d_l$ of the  mention encoder's output (equal to the dimension of the context encoder's ourput), the input dropout keep probability $p_i$ and output dropout keep probability $p_o$ for LSTM layers (in context encoder and LSTM mention encoder), the L2 regularization parameter $\lambda$, the factor of  hierarchical loss normalization $\alpha$ ($\alpha>0$ means use the normalization), BN (whether using Batch normalization), the max step $S_{lp}$ of the label propagation, the max length $S_m$ of Markov chain, the influence parameter $\lambda_{clsc}$ of CLSC, the batch size $B$, the number $n$ of hidden layers in $q$ and the number $h_n$ of hidden units of the hidden layers.  We implement all models using Tensorflow\footnote{The code for experiments is available at https://github. com/herbertchen1/NFETC-CLSC}.

\subsection{Performance comparison and analysis}
Table \ref{tb:results} shows performance comparison between the proposed CLSC model and state-of-the-art FET systems. On both benchmarks, the CLSC model achieves the best performance in all three metrics. When focusing on the comparison between \textbf{NFETC} and CLSC, we have following observation:
\begin{itemize}
    \item Compact Latent Space Clustering shows its effectiveness on both clean data and noisy data. By applying CLSC regularization on the basic \textbf{NFETC} model, we observe consistent and significant performance boost;
    \item Hierarchical-aware loss shows significant advantage on the OntoNotes dataset, while showing insignificant performance boost on the BBN dataset. This is due to different distribution of labels on the test set. The proportion of terminal types of the test set is $69\%$ for the BBN dataset, while is only $33\%$ on the OntoNotes dataset. Thus, applying hierarchical-aware loss on the BBN dataset brings little improvement;
    \item Both algorithms are able to utilize noisy data to improve performance, so we would like to further study their performance in different noisy scenarios in following discussions.
\end{itemize}

\subsection{How robust are the methods to the proportion of noisy data?}

\begin{figure}[!htb]
    \centering
    \subfloat{\label{Fig:R1}
    \hspace{-3em}
    \includegraphics[width=0.9\linewidth]{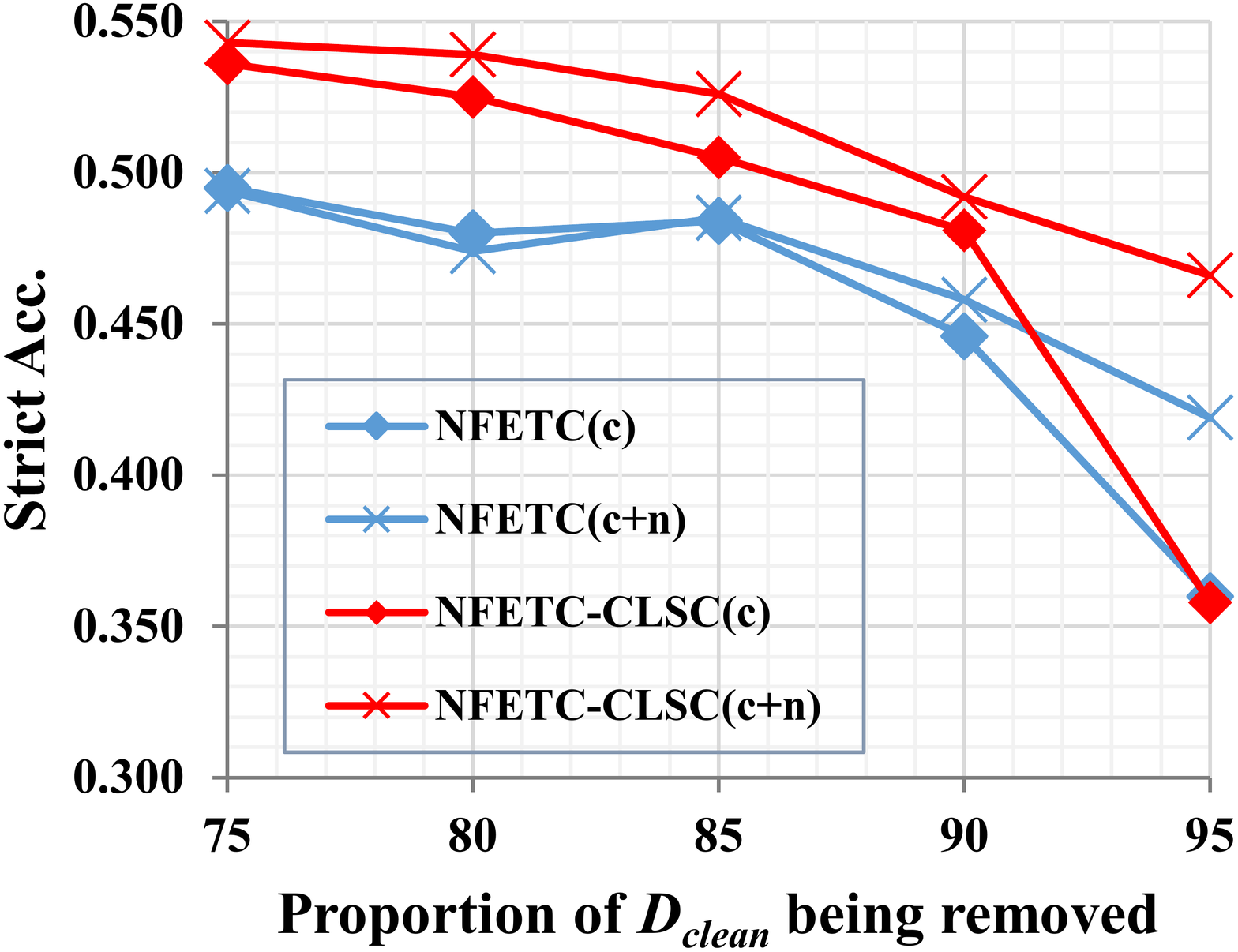}}
    \centering
    \quad
    \subfloat{\label{Fig:R2}
    \hspace{-3em}
    \includegraphics[width=0.9\linewidth]{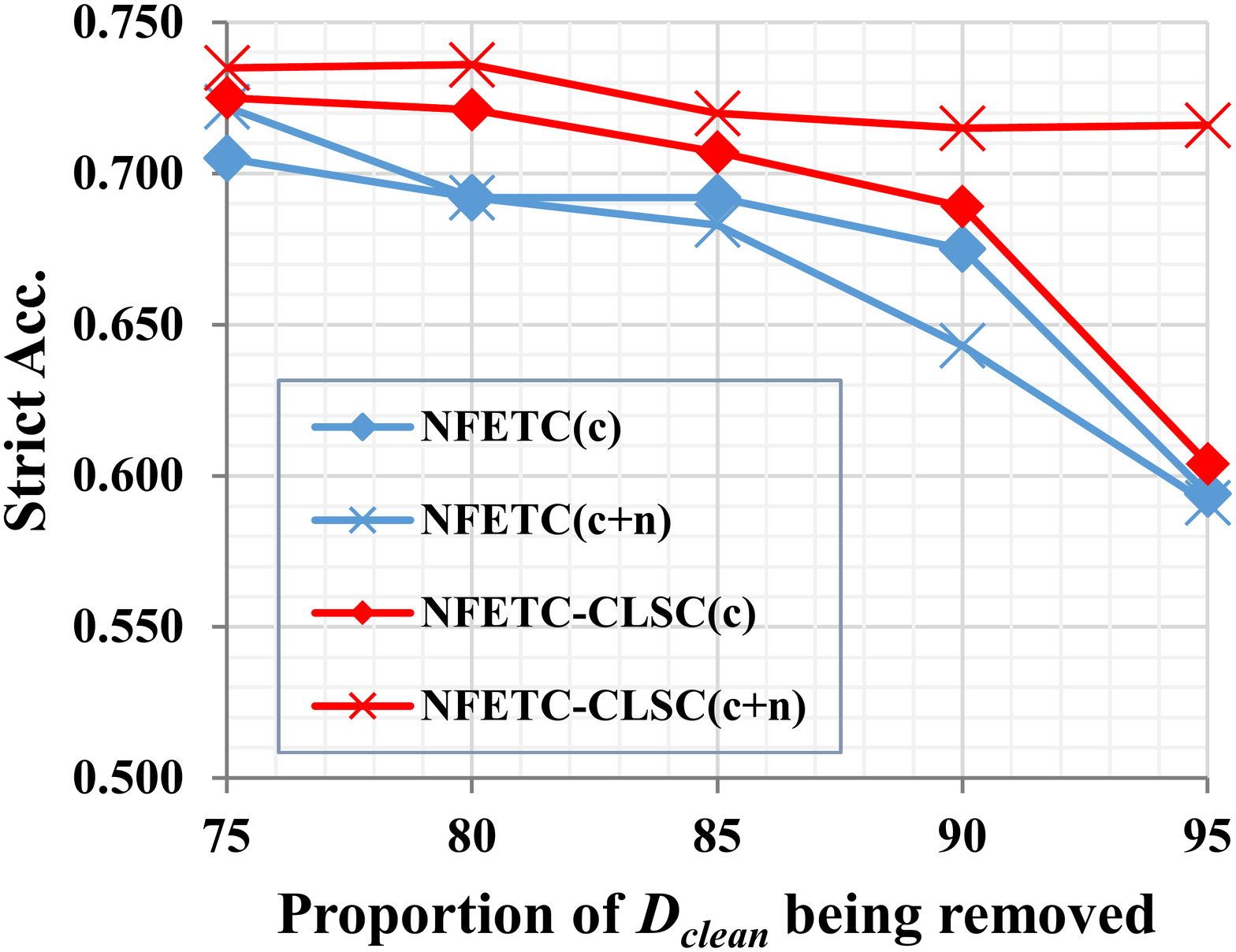}}
    \centering
    \caption{Performance comparison between \textbf{NFETC-CLSC} and \textbf{NFETC} by removing $75\%$-$95\%$ clean data.}
    \label{fig:noise compare}
\end{figure}

By principle, with sufficient amount of clean training data, most typing systems can achieve satisfying performance. To further study the robustness of the methods to label noise, we compare their performance with the presence of $25\%, 20\%, 15\%, 10\%$ and $5\%$ clean training data and all noisy training data. Figure \ref{fig:noise compare} shows the performance curves as the proportion of clean data drops. As it reveals, the CLSC model consistently wins in the comparison. The advantage is especially clear on the BBN dataset, which offers less amount of training data. Note that, with only $27.9\%$ of training data (when only leaving $5\%$ clean data) on the BBN dataset, the CLSC model yield a comparable result with the \textbf{NFETC} model trained on full data. This comparison clearly shows the superiority of our approach in the effectiveness of utilizing noisy data.

\subsection{Ablation: Do Markov Chains improve typing performance?}
Table \ref{tb:l1step-ab} shows the performance of CLSC with one-step transition ($\mathcal{L}_{1-step}$) and with Markov Chains ($\mathcal{L}_{clsc}$) as described in Section \ref{sec:method-compact}. Results show that the use of Markov Chains does bring improvement to the overall performance, which is consistent with the model intuition.

\section{Related Work}
Named entity Recognition (NER) has been excavated for a long time \cite{collins1999unsupervised,manning2014stanford}, which classifies coarse-grained types (e.g. person, location). Recently, \cite{nagesh2018exploration,nagesh2018keep} applied ladder network \cite{rasmus2015semi} to coarse-grained entity classification in a semi-supervised learning fashion.
 \cite{ling2012fine} proposed Fine-Grained Entity Recognition (FET). They used distant supervision to get training corpus for FET. Embedding techniques was applied to learn feature representations since \cite{yogatama2015embedding,dong2015hybrid}. \cite{shimaoka2016attentive} introduced attention mechanism for FET to capture informative words. \cite{xin2018improving} used the TransE entity embeddings \cite{bordes2013translating} as the query vector of attention. \\
 Early works ignore the out-of-context noise, \cite{gillick2014context} proposed context dependent FET and use three heuristics to clean the noisy labels with the side effect of losing training data. To utilize noisy data, \cite{ren2016afet} distinguished the loss function of noisy data from clean data via partial label loss (\textbf{PLL}). \cite{abhishek2017fine,xu2018neural} proposed variants of \textbf{PLL}, which still suffer from confirmation bias. \cite{xu2018neural} proposed hierarchical loss to handle over-specific noise. On top of \textbf{AFET}, \cite{ren2016label} proposed a method \textbf{PLE} to reduce the label noise, which lead to a great success in FET. Because label noise reduction is separated from the learning of FET, there might be error propagation problem. Recently, \cite{xin2018put} proposed utilizing a pretrained language model measures the compatibility between context and type names, and use it to repel the interference of noisy labels. However, the compatibility got by language model may not be right and type information is defined by corpus and annotation guidelines rather than type names as is mentioned in \cite{azad2018unified}.  In addition, there are some work about entity-level typing which aim to figure out the types of entities in KB \cite{yaghoobzadeh2015corpus,jin2018attributed}.
\begin{table}
\centering
\small
\begin{tabular}{|l|c|}
\hline
                  & Strict Acc. \\ \hline
\textbf{CLSC(c)}$(\mathcal{L}_{1-step})$ & 72.0$\pm$0.1    \\ \hline
\textbf{CLSC(c)}$(\mathcal{L}_{clsc})$   & 73.0$\pm$0.3    \\ \hline
\textbf{CLSC(c+n)}$(\mathcal{L}_{1-step})$   & 73.0$\pm$0.1    \\ \hline
\textbf{CLSC(c+n)}$(\mathcal{L}_{clsc})$   & 74.7$\pm$0.3    \\ \hline
\end{tabular}
\caption{The comparison of $\mathcal{L}_{1-step}$ and $\mathcal{L}_{clsc}$ on BBN.}\label{tb:l1step-ab}
\end{table}
\section{Conclusion}
In this paper, we propose a new method for distantly supervised fine-grained entity typing, which leverages imperfect annotations as model regularization via Compact Latent Space Clustering (CLSC). Experiments on two standard benchmarks demonstrate that our  method  consistently  outperforms state-of-the-art models. Further study reveals our method is more robust than the former state-of-the-art approach as the portion of noisy data rises.
The proposed method is general for other tasks with imperfect annotation. As a part of future investigation, 
we plan to apply the approach to other distantly supervised tasks, such as relation extraction.
\section{Acknowledgments}
This work has been supported in part by NSFC (No.61751209, U1611461), Zhejiang University-iFLYTEK Joint Research Center, Chinese Knowledge Center of Engineering Science and Technology (CKCEST), Engineering Research Center of Digital Library, Ministry of Education. Xiang Ren's research has been supported in part by National Science Foundation SMA 18-29268.
\bibliography{naaclhlt2019}

\begin{thebibliography}{36}
\expandafter\ifx\csname natexlab\endcsname\relax\def\natexlab#1{#1}\fi

\bibitem[{Abhishek et~al.(2017)Abhishek, Anand, and Awekar}]{abhishek2017fine}
Abhishek Abhishek, Ashish Anand, and Amit Awekar. 2017.
\newblock Fine-grained entity type classification by jointly learning
  representations and label embeddings.
\newblock In \emph{Proceedings of the 15th Conference of the European Chapter
  of the Association for Computational Linguistics: Volume 1, Long Papers},
  volume~1, pages 797--807.

\bibitem[{Azad et~al.(2018)Azad, Ganesan, Anand, Awekar
  et~al.}]{azad2018unified}
Amar~Prakash Azad, Balaji Ganesan, Ashish Anand, Amit Awekar, et~al. 2018.
\newblock A unified labeling approach by pooling diverse datasets for entity
  typing.
\newblock \emph{arXiv preprint arXiv:1810.08782}.

\bibitem[{Bergstra et~al.(2013)Bergstra, Yamins, and
  Cox}]{bergstra2013hyperopt}
James Bergstra, Dan Yamins, and David~D Cox. 2013.
\newblock Hyperopt: A python library for optimizing the hyperparameters of
  machine learning algorithms.
\newblock In \emph{Proceedings of the 12th Python in Science Conference}, pages
  13--20. Citeseer.

\bibitem[{Bordes et~al.(2013)Bordes, Usunier, Garcia-Duran, Weston, and
  Yakhnenko}]{bordes2013translating}
Antoine Bordes, Nicolas Usunier, Alberto Garcia-Duran, Jason Weston, and Oksana
  Yakhnenko. 2013.
\newblock Translating embeddings for modeling multi-relational data.
\newblock In \emph{Advances in neural information processing systems}, pages
  2787--2795.

\bibitem[{Collins and Singer(1999)}]{collins1999unsupervised}
Michael Collins and Yoram Singer. 1999.
\newblock Unsupervised models for named entity classification.
\newblock In \emph{1999 Joint SIGDAT Conference on Empirical Methods in Natural
  Language Processing and Very Large Corpora}.

\bibitem[{Daiber et~al.(2013)Daiber, Jakob, Hokamp, and
  Mendes}]{daiber2013improving}
Joachim Daiber, Max Jakob, Chris Hokamp, and Pablo~N Mendes. 2013.
\newblock Improving efficiency and accuracy in multilingual entity extraction.
\newblock In \emph{Proceedings of the 9th International Conference on Semantic
  Systems}, pages 121--124. ACM.

\bibitem[{Dong et~al.(2015)Dong, Wei, Sun, Zhou, and Xu}]{dong2015hybrid}
Li~Dong, Furu Wei, Hong Sun, Ming Zhou, and Ke~Xu. 2015.
\newblock A hybrid neural model for type classification of entity mentions.
\newblock In \emph{IJCAI}, pages 1243--1249.

\bibitem[{Dong et~al.(2014)Dong, Gabrilovich, Heitz, Horn, Lao, Murphy,
  Strohmann, Sun, and Zhang}]{dong2014knowledge}
Xin Dong, Evgeniy Gabrilovich, Geremy Heitz, Wilko Horn, Ni~Lao, Kevin Murphy,
  Thomas Strohmann, Shaohua Sun, and Wei Zhang. 2014.
\newblock Knowledge vault: A web-scale approach to probabilistic knowledge
  fusion.
\newblock In \emph{Proceedings of the 20th ACM SIGKDD international conference
  on Knowledge discovery and data mining}, pages 601--610. ACM.

\bibitem[{Gillick et~al.(2014)Gillick, Lazic, Ganchev, Kirchner, and
  Huynh}]{gillick2014context}
Dan Gillick, Nevena Lazic, Kuzman Ganchev, Jesse Kirchner, and David Huynh.
  2014.
\newblock Context-dependent fine-grained entity type tagging.
\newblock \emph{arXiv preprint arXiv:1412.1820}.

\bibitem[{Hochreiter and Schmidhuber(1997)}]{hochreiter1997long}
Sepp Hochreiter and J{\"u}rgen Schmidhuber. 1997.
\newblock Long short-term memory.
\newblock \emph{Neural computation}, 9(8):1735--1780.

\bibitem[{Jin et~al.(2018)Jin, Hou, Li, and Dong}]{jin2018attributed}
Hailong Jin, Lei Hou, Juanzi Li, and Tiansi Dong. 2018.
\newblock Attributed and predictive entity embedding for fine-grained entity
  typing in knowledge bases.
\newblock In \emph{Proceedings of the 27th International Conference on
  Computational Linguistics}, pages 282--292.

\bibitem[{Kamnitsas et~al.(2018)Kamnitsas, Castro, Folgoc, Walker, Tanno,
  Rueckert, Glocker, Criminisi, and Nori}]{pmlr-v80-kamnitsas18a}
Konstantinos Kamnitsas, Daniel Castro, Loic~Le Folgoc, Ian Walker, Ryutaro
  Tanno, Daniel Rueckert, Ben Glocker, Antonio Criminisi, and Aditya Nori.
  2018.
\newblock Semi-supervised learning via compact latent space clustering.
\newblock In \emph{Proceedings of the 35th International Conference on Machine
  Learning}, volume~80 of \emph{Proceedings of Machine Learning Research},
  pages 2459--2468, Stockholmsmässan, Stockholm Sweden. PMLR.

\bibitem[{Laine and Aila(2017)}]{Laine2017iclr}
Samuli Laine and Timo Aila. 2017.
\newblock Temporal ensembling for semi-supervised learning.
\newblock In \emph{Proc. International Conference on Learning Representations
  (ICLR)}.

\bibitem[{Lee et~al.(2006)Lee, Hwang, Oh, Lim, Heo, Lee, Kim, Wang, and
  Jang}]{lee2006fine}
Changki Lee, Yi-Gyu Hwang, Hyo-Jung Oh, Soojong Lim, Jeong Heo, Chung-Hee Lee,
  Hyeon-Jin Kim, Ji-Hyun Wang, and Myung-Gil Jang. 2006.
\newblock Fine-grained named entity recognition using conditional random fields
  for question answering.
\newblock In \emph{Asia Information Retrieval Symposium}, pages 581--587.
  Springer.

\bibitem[{Ling and Weld(2012)}]{ling2012fine}
Xiao Ling and Daniel~S Weld. 2012.
\newblock Fine-grained entity recognition.
\newblock In \emph{AAAI}, volume~12, pages 94--100.

\bibitem[{Manning et~al.(2014)Manning, Surdeanu, Bauer, Finkel, Bethard, and
  McClosky}]{manning2014stanford}
Christopher Manning, Mihai Surdeanu, John Bauer, Jenny Finkel, Steven Bethard,
  and David McClosky. 2014.
\newblock The stanford corenlp natural language processing toolkit.
\newblock In \emph{Proceedings of 52nd annual meeting of the association for
  computational linguistics: system demonstrations}, pages 55--60.

\bibitem[{Mintz et~al.(2009)Mintz, Bills, Snow, and
  Jurafsky}]{mintz2009distant}
Mike Mintz, Steven Bills, Rion Snow, and Dan Jurafsky. 2009.
\newblock Distant supervision for relation extraction without labeled data.
\newblock In \emph{Proceedings of the Joint Conference of the 47th Annual
  Meeting of the ACL and the 4th International Joint Conference on Natural
  Language Processing of the AFNLP: Volume 2-Volume 2}, pages 1003--1011.
  Association for Computational Linguistics.

\bibitem[{Nagesh and Surdeanu(2018{\natexlab{a}})}]{nagesh2018exploration}
Ajay Nagesh and Mihai Surdeanu. 2018{\natexlab{a}}.
\newblock An exploration of three lightly-supervised representation learning
  approaches for named entity classification.
\newblock In \emph{Proceedings of the 27th International Conference on
  Computational Linguistics}, pages 2312--2324.

\bibitem[{Nagesh and Surdeanu(2018{\natexlab{b}})}]{nagesh2018keep}
Ajay Nagesh and Mihai Surdeanu. 2018{\natexlab{b}}.
\newblock Keep your bearings: Lightly-supervised information extraction with
  ladder networks that avoids semantic drift.
\newblock In \emph{Proceedings of the 2018 Conference of the North American
  Chapter of the Association for Computational Linguistics: Human Language
  Technologies, Volume 2 (Short Papers)}, volume~2, pages 352--358.

\bibitem[{Pennington et~al.(2014)Pennington, Socher, and
  Manning}]{pennington2014glove}
Jeffrey Pennington, Richard Socher, and Christopher~D. Manning. 2014.
\newblock \href {http://www.aclweb.org/anthology/D14-1162} {Glove: Global
  vectors for word representation}.
\newblock In \emph{Empirical Methods in Natural Language Processing (EMNLP)},
  pages 1532--1543.

\bibitem[{Raiman and Raiman(2018)}]{DBLP:conf/aaai/RaimanR18}
Jonathan Raiman and Olivier Raiman. 2018.
\newblock \href
  {https://www.aaai.org/ocs/index.php/AAAI/AAAI18/paper/view/17148} {Deeptype:
  Multilingual entity linking by neural type system evolution}.
\newblock In \emph{Proceedings of the Thirty-Second {AAAI} Conference on
  Artificial Intelligence, (AAAI-18), the 30th innovative Applications of
  Artificial Intelligence (IAAI-18), and the 8th {AAAI} Symposium on
  Educational Advances in Artificial Intelligence (EAAI-18), New Orleans,
  Louisiana, USA, February 2-7, 2018}, pages 5406--5413.

\bibitem[{Rasmus et~al.(2015)Rasmus, Berglund, Honkala, Valpola, and
  Raiko}]{rasmus2015semi}
Antti Rasmus, Mathias Berglund, Mikko Honkala, Harri Valpola, and Tapani Raiko.
  2015.
\newblock Semi-supervised learning with ladder networks.
\newblock In \emph{Advances in Neural Information Processing Systems}, pages
  3546--3554.

\bibitem[{Ren et~al.(2016{\natexlab{a}})Ren, He, Qu, Huang, Ji, and
  Han}]{ren2016afet}
Xiang Ren, Wenqi He, Meng Qu, Lifu Huang, Heng Ji, and Jiawei Han.
  2016{\natexlab{a}}.
\newblock Afet: Automatic fine-grained entity typing by hierarchical
  partial-label embedding.
\newblock In \emph{Proceedings of the 2016 Conference on Empirical Methods in
  Natural Language Processing}, pages 1369--1378.

\bibitem[{Ren et~al.(2016{\natexlab{b}})Ren, He, Qu, Voss, Ji, and
  Han}]{ren2016label}
Xiang Ren, Wenqi He, Meng Qu, Clare~R Voss, Heng Ji, and Jiawei Han.
  2016{\natexlab{b}}.
\newblock Label noise reduction in entity typing by heterogeneous partial-label
  embedding.
\newblock In \emph{Proceedings of the 22nd ACM SIGKDD international conference
  on Knowledge discovery and data mining}, pages 1825--1834. ACM.

\bibitem[{Shimaoka et~al.(2016)Shimaoka, Stenetorp, Inui, and
  Riedel}]{shimaoka2016attentive}
Sonse Shimaoka, Pontus Stenetorp, Kentaro Inui, and Sebastian Riedel. 2016.
\newblock An attentive neural architecture for fine-grained entity type
  classification.
\newblock In \emph{Proceedings of the 5th Workshop on Automated Knowledge Base
  Construction}, pages 69--74.

\bibitem[{Tarvainen and Valpola(2017)}]{tarvainen2017mean}
Antti Tarvainen and Harri Valpola. 2017.
\newblock Mean teachers are better role models: Weight-averaged consistency
  targets improve semi-supervised deep learning results.
\newblock In \emph{Advances in neural information processing systems}, pages
  1195--1204.

\bibitem[{Vaswani et~al.(2017)Vaswani, Shazeer, Parmar, Uszkoreit, Jones,
  Gomez, Kaiser, and Polosukhin}]{vaswani2017attention}
Ashish Vaswani, Noam Shazeer, Niki Parmar, Jakob Uszkoreit, Llion Jones,
  Aidan~N Gomez, {\L}ukasz Kaiser, and Illia Polosukhin. 2017.
\newblock Attention is all you need.
\newblock In \emph{Advances in Neural Information Processing Systems}, pages
  5998--6008.

\bibitem[{Weischedel and Brunstein(2005)}]{weischedel2005bbn}
Ralph Weischedel and Ada Brunstein. 2005.
\newblock Bbn pronoun coreference and entity type corpus.
\newblock \emph{Linguistic Data Consortium, Philadelphia}, 112.

\bibitem[{Weischedel et~al.(2013)Weischedel, Palmer, Marcus, Hovy, Pradhan,
  Ramshaw, Xue, Taylor, Kaufman, Franchini et~al.}]{weischedel2013ontonotes}
Ralph Weischedel, Martha Palmer, Mitchell Marcus, Eduard Hovy, Sameer Pradhan,
  Lance Ramshaw, Nianwen Xue, Ann Taylor, Jeff Kaufman, Michelle Franchini,
  et~al. 2013.
\newblock Ontonotes release 5.0 ldc2013t19.
\newblock \emph{Linguistic Data Consortium, Philadelphia, PA}.

\bibitem[{Xin et~al.(2018{\natexlab{a}})Xin, Lin, Liu, and
  Sun}]{xin2018improving}
Ji~Xin, Yankai Lin, Zhiyuan Liu, and Maosong Sun. 2018{\natexlab{a}}.
\newblock Improving neural fine-grained entity typing with knowledge attention.

\bibitem[{Xin et~al.(2018{\natexlab{b}})Xin, Zhu, Han, Liu, and
  Sun}]{xin2018put}
Ji~Xin, Hao Zhu, Xu~Han, Zhiyuan Liu, and Maosong Sun. 2018{\natexlab{b}}.
\newblock Put it back: Entity typing with language model enhancement.
\newblock In \emph{Proceedings of the 2018 Conference on Empirical Methods in
  Natural Language Processing}, pages 993--998.

\bibitem[{Xu and Barbosa(2018)}]{xu2018neural}
Peng Xu and Denilson Barbosa. 2018.
\newblock Neural fine-grained entity type classification with hierarchy-aware
  loss.
\newblock In \emph{Proceedings of the 2018 Conference of the North American
  Chapter of the Association for Computational Linguistics: Human Language
  Technologies, Volume 1 (Long Papers)}, volume~1, pages 16--25.

\bibitem[{Yaghoobzadeh and Sch{\"u}tze(2015)}]{yaghoobzadeh2015corpus}
Yadollah Yaghoobzadeh and Hinrich Sch{\"u}tze. 2015.
\newblock Corpus-level fine-grained entity typing using contextual information.
\newblock In \emph{Proceedings of the 2015 Conference on Empirical Methods in
  Natural Language Processing}, pages 715--725.

\bibitem[{Yogatama et~al.(2015)Yogatama, Gillick, and
  Lazic}]{yogatama2015embedding}
Dani Yogatama, Daniel Gillick, and Nevena Lazic. 2015.
\newblock Embedding methods for fine grained entity type classification.
\newblock In \emph{Proceedings of the 53rd Annual Meeting of the Association
  for Computational Linguistics and the 7th International Joint Conference on
  Natural Language Processing (Volume 2: Short Papers)}, volume~2, pages
  291--296.

\bibitem[{Zeng et~al.(2014)Zeng, Liu, Lai, Zhou, and Zhao}]{zeng2014relation}
Daojian Zeng, Kang Liu, Siwei Lai, Guangyou Zhou, and Jun Zhao. 2014.
\newblock Relation classification via convolutional deep neural network.
\newblock In \emph{Proceedings of COLING 2014, the 25th International
  Conference on Computational Linguistics: Technical Papers}, pages 2335--2344.

\bibitem[{Zhu and Ghahramani(2002)}]{zhu2002learning}
Xiaojin Zhu and Zoubin Ghahramani. 2002.
\newblock Learning from labeled and unlabeled data with label propagation.

\end{thebibliography}
\bibliographystyle{acl_natbib}
\appendix
\makeatletter
\setlength{\@fptop}{5pt}
\makeatother
\textbf{Appendix}
\section{Hyper parameters}
\begin{table}[t]
\centering
\begin{tabular}{|l|r|r|r|}
\hline
\textbf{}        & \multicolumn{1}{l|}{\textbf{Ont.(C)}} & \multicolumn{1}{l|}{\textbf{BBN(C)}} & \multicolumn{1}{l|}{\textbf{BBN(N)}} \\ \hline
$lr$             & 0.0006                       & 0.0007                               & 0.0007                               \\ \hline
$d_p$            & 70                          & 40                                   & 20                                   \\ \hline
$d_l$            & 1000                          & 1000                                 & 240                                  \\ \hline
$p_i$            & 0.7                          & 0.3                                 & 0.5                                  \\ \hline
$p_o$            & 0.6                          & 1.0                                  & 0.4                                  \\ \hline
$\lambda$        & 0.0000                       & 0.0000                               & 0.0002                               \\ \hline
$\alpha$         & 0.25/0.0                      & 0.4/0.0                              & 0.4/0.0                              \\ \hline
BN               & FALSE                         & FALSE                                & TRUE                                 \\ \hline
$S_{lp}$         & 200                           & 200                                  & -                                    \\ \hline
$S_m$            & 8                           & 12                                   & -                                    \\ \hline
$\lambda_{clsc}$ & 2.0                          & 1.5                                  & -                                    \\ \hline
$B$              & 512                          & 512                                  & 512                                  \\ \hline
$n$            & 2                           & 1                                   & -                                    \\ \hline
$h_n$            & 700                           & 560                                   & -                                    \\ \hline
\end{tabular}
\caption{Hyper parameters of our experiments: (C) denotes CLSC, (N) denotes the hyper parameter is used for NFETC.}\label{tb:hyperpar}
\end{table}

\end{document}